\def\year{2021}\relax
\documentclass[letterpaper]{article} 
\usepackage{aaai21}  
\usepackage{times}  
\usepackage{helvet}  
\usepackage{courier}  
\usepackage{url}  
\usepackage{graphicx}  
\usepackage{natbib}
\usepackage{amsmath,amssymb,amsfonts}
\usepackage{algorithmic}
\usepackage[utf8]{inputenc} 
\usepackage[T1]{fontenc}    
\usepackage{url}            
\usepackage{booktabs}       
\usepackage{nicefrac}       
\usepackage{bm}
\usepackage[algo2e,ruled,linesnumbered]{algorithm2e}
\usepackage[switch]{lineno}
\usepackage{subcaption}
\usepackage{verbatim}

\usepackage{caption} 

\DeclareMathOperator*{\argmin}{argmin}

\newtheorem{Remark}{Remark}

\newcommand{\matA}{\mathbf{A}}
\newcommand{\matB}{\mathbf{B}}
\newcommand{\matC}{\mathbf{C}}
\newcommand{\matD}{\mathbf{D}}
\newcommand{\matE}{\mathbf{E}}

\newcommand{\matG}{\mathbf{G}}

\newcommand{\matM}{\mathbf{M}}
\newcommand{\matN}{\mathbf{N}}
\newcommand{\matO}{\mathbf{O}}
\newcommand{\matP}{\mathbf{P}}

\newcommand{\matU}{\mathbf{U}}
\newcommand{\matV}{\mathbf{V}}
\newcommand{\matW}{\mathbf{W}}
\newcommand{\matX}{\mathbf{X}}
\newcommand{\matY}{\mathbf{Y}}
\newcommand{\matZ}{\mathbf{Z}}
\newcommand{\matI}{\mathbf{I}}
\newcommand{\mattr}{\mathbf{\operatorname{Tr}}}

\newcommand{\one}{\bm{1}}
\newcommand{\balpha}{\mathlarger{\bm{\alpha}}}
\newcommand{\bbeta}{\mathlarger{\bm{\beta}}}
\newcommand{\bgamma}{\mathlarger{\bm{\gamma}}}

\newcommand{\bnabla}{\mathlarger{\bm{\nabla}}}
\newcommand{\bsigma}{\mathlarger{\bm{\sigma}}}
\newcommand{\bomega}{\mathlarger{\bm{\omega}}}

\newcommand{\calD}{\mathcal{D}}
\newcommand{\calE}{\mathcal{E}}
\newcommand{\calF}{\mathcal{F}}
\newcommand{\calG}{\mathcal{G}}

\newcommand{\calP}{\mathcal{P}}
\newcommand{\calV}{\mathcal{V}}
\newcommand{\calX}{\mathcal{X}}
\newcommand{\calY}{\mathcal{Y}}

\title{Permutation-Invariant Subgraph Discovery}
\author{Raghvendra Mall, Shameem A. Puthiya Parambath, Ting Yu, Sanjay Chawla\\
Qatar Computing Research Institute, Hamad Bin Khalifa University, Doha, Qatar \\
and \\
Han Yufei \\
INRIA\\
}

\begin{document}
\maketitle

\begin{abstract}
We introduce Permutation and Structured Perturbation Inference (PSPI), a new problem formulation which abstracts many graph matching tasks that arise in systems biology. PSPI can be viewed as a robust formulation of the permutation inference or graph matching, where the objective is to find a permutation between two graphs under the assumption that a set of edges may have undergone a perturbation due to an underlying cause. For example, suppose there are two gene regulatory networks $\matX$ and $\matY$ from a diseased and normal tissue respectively. Then, the PSPI problem can be used to detect if there has been a structural change between the two networks which can serve as a signature of the disease. 
Besides the new problem formulation, we propose an ADMM algorithm ({\tt STEPD}) to solve a relaxed version of the  PSPI problem.
An extensive case study on comparative gene regulatory networks (GRNs) is used to demonstrate that {\tt STEPD} is able to accurately infer structured perturbations and thus provides a tool for computational biologists to identify novel prognostic signatures. A spectral analysis confirms that {\tt STEPD} can recover small clique-like perturbations making it a useful tool for detecting permutation-invariant changes in graphs. 
\end{abstract}

\maketitle

\section{Introduction}
Given two graphs $\matX$ and $\matY$ defined on a common vertex set $\calV$, can
we \emph{infer} a permutation matrix $\matP$ and a  perturbation matrix $\matZ$ s.t. $
\matY \sim \matP\matX\matP' + \matZ $.

We refer to the above as the PSPI (Permutation and Structured Perturbation Inference)
problem, which can be considered analogous to the famous robust principal
component analysis problem
introduced in \cite{candes2011robust}, where the objective is to decompose a rectangular matrix $\mathbf{L}$ into a sum of a low rank matrix $\mathbf{M}$ and a 
structured sparse matrix $\mathbf{S}$.  However the PSPI problem is substantially harder due to the combinatorial nature of the search over permutation matrices. There are $n!$ permutations on a graph of size $n$.

PSPI is a new problem and abstracts many practical tasks. While our primary motivation is from cancer research, PSPI can be
applied in many network analysis scenarios. To understand the nature of cancer, scientists often compare the gene regulatory networks 
of healthy ($\matX$) and diseased samples ($\matY$) \cite{mall2018rgbm}. Studies have shown that regulatory networks undergo some amount of localized re-wirings as cancer progresses. Given two such regulatory networks, it is important to detect not only the correspondence between the nodes of the networks, but also structured localized perturbations~\cite{zhang2005general,tiffany,mall2017detection}. Even though gene networks are vertex annotated, genes are 
known to take over the functionality of other genes making a direct difference comparison between two networks often misleading. Thus, detecting perturbations in a graph without taking permutation into account is fundamentally a mis-identification problem.

Another example can be found in neurology research. $\matX$ and $\matY$ represent two networks over the same set of neurons from healthy and diseased brain tissue respectively. It was shown in  \cite{demarin2016arts} that neurons in the brain could ``reorganize'' after a traumatic event and take over its role allowing normal functioning of an individual. Usually, the combinations of genes and neurons that provide specific biological functions have relatively fixed combinatorial patterns, compared to random mutation. These combination patterns form structured cliques in the graphs. Matching the graphs of genes and neurons helps identify these structured and localized changes for downstream research.





\begin{figure}[!ht]
\centering
\includegraphics[width=0.14\textwidth]{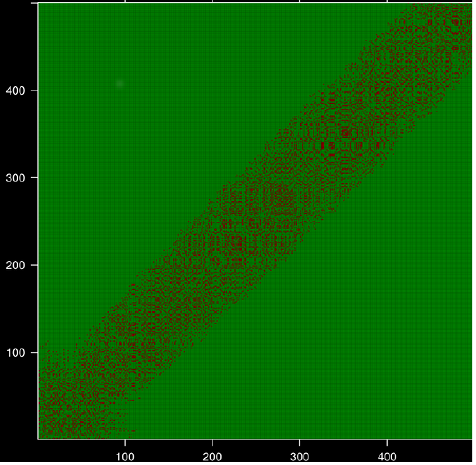}
\includegraphics[width=0.14\textwidth]{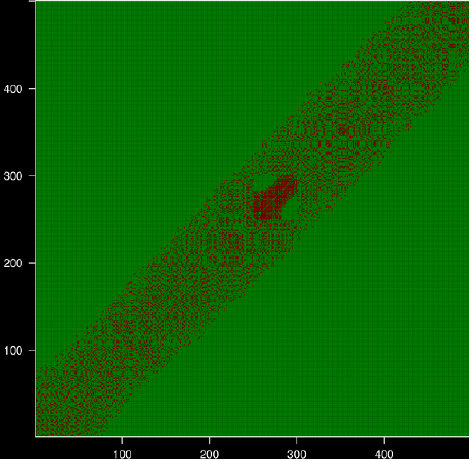} 
\includegraphics[width=0.16\textwidth]{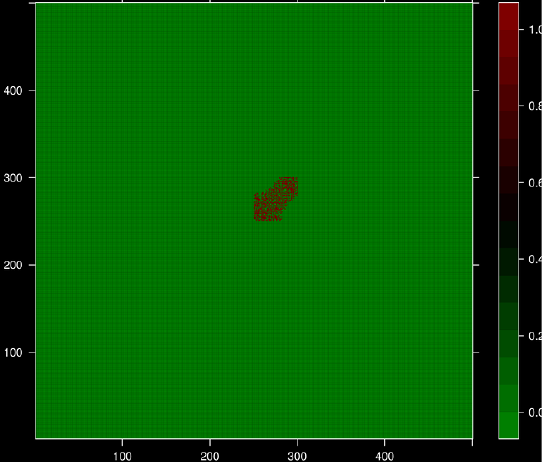}
\caption{The left plot shows the original random geometric graph $\matX$ and middle plot depicts the isomorphic adjacency matrix with random noise, permutation (first 50 nodes) and structured perturbation $\matY$. The perturbation corresponds to structured rewiring in $\matX$. Our STEPD approach can correctly infer the perturbation matrix corresponding to $\hat{\matZ}$. }
\label{fig:toy_image}
\vspace{-1mm}
\end{figure}

\indent {\bf Example:} Consider the toy example given in Fig~\ref{fig:toy_image}.
There is an adjacency matrix ($\matX$) of a random geometric graph of $500$ nodes (leftmost) and a permuted and perturbed version of it ($\matY$). The first $50$ nodes in the isomorphic $\matY$ are  permuted  w.r.t. $\matX$ and it additionally  contains some random noise (in practice this happens due to noise in the acquisition process). The goal of PSPI is: Given the paired networks, we infer the permutation matrix $\matP$ and the structured perturbation matrix $\matZ$. The rightmost image in Fig \ref{fig:toy_image} shows the inferred $\matZ$ which is referred as $\hat{\matZ}$.

The problem of PSPI in paired networks is closely related to 
graph matching. In fact, the permutation inference corresponds exactly to the graph matching problem \cite{conte2004thirty}.
In graph matching, one is interested in finding the correspondence (isomorphism) between the nodes of two graphs such that the graphs are `structurally the same'.
In practice, the observed graphs are subject to random noise due to many factors including the noise in the acquisition process and one is interested in approximate matching.
In general, graph matching is a difficult combinatorial problem, and the complexity class of the corresponding decision problem is not yet known. It is worth noting that computer vision is another domain where graph matching is a popular topic \cite{ChoECCV2010,DymNadACMTrans2017,LeHuuCVPR2017,WangPAMI2018}. 
They focus on recovering the node correspondences between object components of different scene scenarios. These works do not consider the impact of noise corruption \cite{ChoECCV2010,DymNadACMTrans2017,LeHuuCVPR2017,WangPAMI2018}, or only assume the existence of random and sparse edge addition/deletion \cite{YanICCV2015,KuiJiaIJCV2016}. In contrast, our study focuses on identifying and reconstructing \emph{structured perturbations implanted in graphs}. 
The PSPI problem which we investigate here is significantly different and to the best our knowledge this is the first attempt to solve this problem.

Our major contributions are the following:
\begin{enumerate}
\item
We introduce  a new problem Permutation and Structure Perturbation Inference (PSPI), for inferring both permutations and structured perturbations in paired networks.
\item  We formulate PSPI as a non-linear integer program whose relaxation turns out to be a bi-convex program and propose an algorithm termed as STructurEd Perturbation and permutation Detection (STEPD) based on ADMM principles to solve the bi-convex program.
\item We show STEPD performs better than several state-of-the-art graph matching solutions on several simulated and benchmark Scale-Free and Erdos-R$\grave{e}$nyi networks.
\item We demonstrate that STEPD method can identify biologically relevant structural differences between real-world networks of healthy and tumor patients.
\end{enumerate}

\section{Related Work}
Most of the prior work related to the PSPI problem is confined to the permutation inference problem in graphs i.e. graph matching, which has been extensively studied within the computer vision and bioinformatics. 
Due to the combinatorial nature of the problem, exact methods based on full or partial enumeration do not scale well, and approximate algorithms are used in practice.
We briefly discuss some of the approximation algorithms closely related to our work. A good review of this topic can be found in \cite{conte2004thirty}.

In \cite{umeyama1988eigendecomposition}, one of the early work on graph matching, the authors proposed an algorithm based on spectral decomposition of the adjacency matrices.
The core idea 
lies in representing the nodes as the orthogonal eigenvectors of the adjacency matrices in the eigenspace, the vector space spanned by the eigenvectors.
Then the problem of finding the optimal matching 
reduces to the problem of finding the permutation matrix.
In \cite{umeyama1988eigendecomposition}, authors proposed to find the permutation matrix by optimizing the euclidean distance between the rows of the absolute value orthogonal eigenvectors using techniques like the Hungarian method. In \citep{singh2007pairwise}, the authors proposed the IsoRank algorithm, similar to PageRank, to find the maximum common subgraph between two protein networks by associating a topological similarity score to the nodes, and solving the eigenvalue problem using the power method.

One of the most commonly used techniques to solve the graph matching problem approximately, including this work,
is based on the relaxation of a problem specific discrete objective function.
The original combinatorial objective function is relaxed to a tractable optimization problem which can be efficiently solved.
Often the element-wise $\ell_1$ or $\ell_2$ (Frobenius) norm of the matching error, defined as the number of the adjacency disagreements between the two graphs i.e. $\|\matP\matX-\matY\matP\|_p, p>0$, is used as the objective function.
In \cite{zaslavskiy2009path}, the authors proposed two algorithms based on convex-concave relaxations. The convex relaxation based algorithm called QCV is obtained by relaxing the Frobenius norm of the matching error to the convex set of double stochastic matrix.
In the second PATH algorithm, a non-linear objective function is formed by linearly combining the convex relaxation term and a concave relaxation term.
The final objective function is solved using the conditional gradient method.
Recently, in \citep{fiori2013robust} an algorithm was proposed for multi-modal graph matching (MGM) based on convex relaxation of the Frobenius norm of the matching error. A relaxed version of the problem was solved in \citep{vogelstein2015fast} using projected gradient descent and conditional gradient in the context of comparing brain images. Other graph matching methods include relaxation labeling, replicator equations and tree search etc. Interested readers are referred to the survey paper by \citep{conte2004thirty} and the references therein.
Our proposed algorithm differs from the above discussed approaches, as we are interested in inferring both structured perturbation and permutations in paired networks.

Another relevant topic is graph classification. Previous works \cite{Przulj2006eccb,Shervashidze2010nips,Kashima2003icml,Borgwardt2005ICDM,Bryan2014KDD,Yanardag2015KDD,Grover2016KDD,Hamilton2017NIPS,Lee2018KDD,kipf2016semisupervised} focused on deciding whether two graphs contain components, such as sub-graphs, of similar 
structures with random-walk based graph kernels or various neural network architectures. The output is a binary classification decision. By contrast, the PSPI problem focuses on more fine-grained information: it aims at recovering the clique based perturbation and permutation, rather than producing a binary classification output. 

\section{Problem Description}
\label{sec:prob_def}

Given two graphs of size $n$, $\calX$ and $\calY$, we denote their adjacency matrices as $\matX$ and $\matY$ respectively. We assume that the graphs are undirected with no self-loops and parallel edges. 
Mathematically, the graph matching problem can be stated as finding a permutation matrix $\matP$ (bijection between nodes), such that the two graphs are `edge preserving' isomorphic i.e. $\matY = \matP\matX\matP'$.
In practice, the observed graphs are subject to random noise and might not be perfectly isomorphic. Formally, we can write the problem as $\matY = \matP\matX\matP' + \matE$ where $\matE$ represents the random noise.
In a more realistic settings, in graph matching one is interested in finding a permutation matrix which is `edge preserving' in an `optimal sense'.
The optimality is usually defined in terms of the matching error which is defined as the number of adjacency disagreements between $\matX$ and $\matY$ quantified using the Frobenius matrix norm.
We use $\|\cdot\|$ to indicate matrix Frobenius norm.
Thus the objective function becomes
    ${\displaystyle \argmin_{\matP \in \calP} \|\matY-\matP\matX\matP'\|^2}$ where $\calP$ is the set of all permutation matrices.
In general, the combinatorial nature of the permutation matrix search makes the problem NP.
A common strategy is to relax the $\calP$ to a compact set.
The convex hull of the permutation matrix corresponds to the set of double stochastic matrix, also called Birkhoff polytope, and the original problem reduces to the relaxed problem
    ${\displaystyle \argmin_{\matP \in \calD} \|\matY-\matP\matX\matP'\|^2}$
where $\calD$ is the Birkhoff polytope i.e. $\calD = \{\matA \in \Re^{n\times n}, \matA \geq 0, \matA\one = \one, \matA'\one = \one\}$, $\one$ being the $n$-dimensional vector of 1s.

As stated  in the introduction, in many real world networks, in addition to the random noise represented by  spurious edges, we have localized structured perturbations which indicates `re-wirings` or new activity in specific regions of the network.
Given two graphs in the form of adjacency matrices $\matX$ and $\matY$, we aim to find the optimal matching between the graphs and infer only the localized structured perturbations.
We propose a simple extension to the original graph matching problem, as 
    ${\displaystyle \matY = \matP\matX\matP' + \matZ + \matE}$.
Here, $\matZ$ captures the localized structured perturbations and can be conceived as the adjacency matrix of a difference graph and hence symmetric.
We limit ourselves to the case where the structured perturbation consists of only addition of new clusters of edges but not the removal.
Hence, $\matZ$ is a symmetric binary matrix.
The binary restriction on $\matZ$ makes the problem combinatorial in $\matZ$, hence we relax $\matZ$ to the set of matrices such that entries are between $[0,1]$.

\subsection{Optimization Framework}
Following \citep{zaslavskiy2009path,fiori2013robust}, we propose an optimization framework as problem solution.
In case of $\matP$, similar to \citep{fiori2013robust}, we use an objective function which disregards the spurious mismatching edges and looks for group structure between $\matP\matX$ and $\matY\matP$ i.e. the objective function encourages that $\matP\matX$ and $\matY\matP$ to share the same active set except for spurious edges. We enforce a group regularization on $\matZ$ to capture any localized group structure present in $\matZ$. We also assume that $\matZ$ is a very sparse matrix, as the perturbation occur at very specific regions of the network.
Thus, our objective function for $\matZ$ takes into account the two aspects \emph{(i)} perturbations are localized but structured and \emph{(ii)} perturbation appears only at a small number of regions of the network.

Combining the different objectives, our final objective function becomes,
\small
\begin{multline}
    \label{eq:pspi_obj}
    \argmin_{\matP\in\calD,\matZ}  \sum_{i,j} \big\|(\balpha_{ij},\bbeta_{ij})\big\|_2 + \|\matP\matX + \matZ\matP - \matY\matP\|^2 + \nu\|\matZ\|_1  \\
~~~~~~~~+ \mu\sum_{i}\|\matZ_{i}\|_2 \text{~where~} \balpha = \matP\matX,~~ \bbeta = \matY\matP
\end{multline}
\normalsize
where $\nu$ and $\mu$ are regularization co-efficients.

The group lasso on the pair $(\balpha,\bbeta)$ encourages non-random group of edges to be active, in particular the minimizer of the objective function (\ref{eq:pspi_obj}) with no perturbation is exactly $\matP\matX = \matY\matP$. 
The group lasso on $\matZ$ promotes only localized structured perturbation to be active and the $\ell_1$ regularization on $\matZ$ encourages sparsity.

The optimization problem in Equation~\eqref{eq:pspi_obj} is a bi-convex problem.
The problem is convex in $\matP$, keeping $\matZ$ constant and convex in $\matZ$ keeping $\matP$ constant.
Since, in general bi-convex problems do not have closed-form solutions, iterative methods like alternating direction methods are used in practice.
In the next section, we propose an algorithm to solve the optimization problem given in \eqref{eq:pspi_obj}.

\subsection{ADMM formulation}
Alternating Direction Method of Multipliers is an iterative method to solve non-linear optimization problems in a Gauss-Seidel fashion \cite{boyd2011distributed}, and naturally fits in our optimization framework.
Moreover, ADMM based algorithm is more appealing as the intermediate updates in ADMM scheme naturally lend to a parallel implementation.
Here, we propose an ADMM based algorithm to solve \eqref{eq:pspi_obj}.

Introducing auxiliary variables for $\matC = \matZ$ the augmented Lagrangian for \eqref{eq:pspi_obj} becomes:
\small
\begin{align*}
\begin{split}
\argmin_{\matP\in\calD,\matZ} & \sum_{i,j} \big\|(\balpha_{ij},\bbeta_{ij})\big\|_2 + \frac{1}{2}\|\matP\matX - \matY\matP + \matZ\matP \|_2^2 + \nu\|\matC\|_1\\
&~~~~ + \mu\sum_{i}\|\matC_{i}\|_2 + \frac{\rho}{2}\|\matC-\matZ+\matD\|_2^2 + \\
&~~~~~~~\frac{\rho}{2}\|\balpha-\matP\matX+\matU\|^2 + \frac{\rho}{2}\|\bbeta-\matY\matP+\matV\|^2
\end{split}
\end{align*}
\normalsize
here $\matU, \matV$ and $\matD$ are related to the Lagrange multiplier, and $\rho$ is the penalization parameter.

In ADMM, we iteratively update each variable, keeping others constant, starting with primal variables $\balpha, \bbeta, \matC, \matP, \matZ$ and followed by the dual variables $\matU, \matV$ and $\matD$.
The complete ADMM based algorithm is given in Algorithm~\ref{alg:admm_matching}.
\SetAlFnt{\small}
\begin{algorithm2e}[tb!]
   \caption{ADMM Based Algorithm}
\label{alg:admm_matching}
    \SetKwData{Left}{left}\SetKwData{This}{this}\SetKwData{Up}{up}
    \SetKwFunction{Union}{Union}\SetKwFunction{FindCompress}{FindCompress}
    \SetKwInOut{Input}{Input}\SetKwInOut{Output}{Output}
    \Input {Adjacency matrix $\matX$ for $\matG_1$ and $\matY$ for $\matG_2$, penalty parameter $\rho$}
    Initialize $\balpha = 0, \bbeta = 0, \bgamma = 0,\matZ = 0, \matP = \frac{1}{n}\one'\one,  \matU = 0, \matV = 0, \matW = 0$ \;
    \Repeat { stopping criterion is not satisfied } {
        $(\balpha,\bbeta) = \argmin_{\balpha,\bbeta}~\sum_{i,j}\big\|(\balpha_{ij},\bbeta_{ij}\big\|_2 + $
            \hangindent=5\skiptext\hangafter=1$\frac{\rho}{2}\big\|\balpha-\matP\matX+\matU\big\|_2^2 + \frac{\rho}{2}\big\|\bbeta-\matY\matP+\matV\big\|_2^2$\; \label{alg:1line}
        $\matC = \argmin_{\matC} ~\nu\big\|\matC\big\|_1 + \mu\sum_{i}\big\|\matC_{i}\big\|_2 + $
            \hangindent=15\skiptext\hangafter=1$ \frac{\rho}{2}\big\|\matC-\matZ+\matD\big\|_2^2$\; \label{alg:2line}
        $\matP = \argmin_{\matP \in \calD} \frac{1}{2}\big\|\matP\matX + \matZ\matP - \matY\matP\big\|_2^2 +$
            \hangindent=5\skiptext\hangafter=1$\frac{\rho}{2}\big\|\balpha-\matP\matX+\matU\big\|_2^2 + \frac{\rho}{2}\big\|\bbeta-\matY\matP+\matV\big\|_2^2$\; \label{alg:3line}
        $\matZ = \argmin_{\matZ} \frac{1}{2}\big\|\matP\matX + \matZ\matP - \matY\matP\big\|_2^2 +$
        \hangindent=15\skiptext\hangafter=1$\frac{\rho}{2}\big\|\matC-\matZ+\matD\big\|_2^2$\; \label{alg:4line}
        $\matU = \matU + \balpha-\matP\matX$\;
        $\matV = \matV + \bbeta-\matY\matP$\;
        $\matD = \matD + \matC - \matZ$\;
    }
    \Output{$\matP, \matZ$}
\end{algorithm2e}
We now take a closer look at each of these update steps, and propose efficient methods to solve each of the sub-optimization problems, where the closed-form solution is not available.
We refer to the numbering in Algorithm~\ref{alg:admm_matching} when the sub-optimization problems are referred using numbers.
The subproblem at line \ref{alg:1line} is a well studied group lasso problem, and in \cite{yuan2006model}, authors propose a closed-form solution based on the soft-thresholding operator as given by
    ${\displaystyle S(a,\rho) = \Big[1-\frac{\rho}{\|a\|_2}\Big]_+a}$.
In our settings, $a$ corresponds to the 2-dimensional vector of individual elements of $\matP\matX$ and $\matY\matP$.
Similarly, the solution for the subproblem at line \ref{alg:2line} can be expressed in a closed-form solution.
Similar to the problem at line \ref{alg:1line}, this problem is a group lasso with an additional lasso term on the variable matrix.
As shown in \citep{ming2014stock}, the problem can be independently solved for each column (or for each row due to symmetrization) of $\matC$, and the closed-form solution can be written in terms of the soft-thresholding operator as:
    ${\displaystyle S(a,\mu,\nu) = \Big[\frac{\|a\|_2-\mu}{\nu\|a\|_2}\Big]_+a}$.
Here $a$ represents the column vector of $\matC$.

\subsubsection{Solving for $\matP$}
The subproblem at line \ref{alg:3line} is a constrained convex optimization problem, where the domain is constrained to be in the set of double stochastic matrix (Birkhoff polytope).
Unfortunately, no closed-form solution exists for this problem. Due to the high computation cost of interior-point methods, in practice first-order methods are preferred to solve such problems.
The projected gradient descent algorithm \cite{luenberger2015linear} is a popular technique due to its guaranteed linear convergence rate.
But the projection of $\matP$ to the set of double stochastic matrix does not have a closed-form solution either.
Hence employing gradient based methods will not be computationally efficient.
To solve for $\matP$, we use linearized version of the alternating direction method (ADM), recently proposed in \citep{lin2011linearized}.
Linearized ADM approach is extremely useful for solving ADMM subproblems where a closed-form solution does not exist.
Interestingly, linearization makes the auxiliary variables unnecessary, and hence there is no need to update them.
Moreover, Linearized ADM converges faster than the traditional ADMM procedure \cite{lin2011linearized}.

Linear ADM is a first order method, where we augment the first order Taylor approximation of the objective function with the proximal operator.
Its primary advantage is that if the objective function is quadratic, the augmented objective takes a simple projection on the domain constraint ($\calD$ in our case), thus avoiding the iterative gradient updates.
We now give the update formula for $\matP$, by analyzing the three terms appearing in the optimization problem at line \ref{alg:3line} independently.
By introducing auxiliary variables $\bsigma$ and $\bomega$, for $\matP\matX$ and $\matY\matP$,
the proximal augmented first order Taylor approximation at iteration $j+1$ of ADMM procedure becomes,
\small
\begin{multline*}
\matP_{j+1} = \argmin_{\matP \in \calD} \frac{1}{2}\|\bsigma-\bomega+\matZ\matP_j\|^2 + \langle(\bsigma-\bomega+\matZ\matP_j),
\\
\matP-\matP_j\rangle + \frac{\eta}{2}\|\matP-\matP_j\|^2
\end{multline*}
\normalsize
where $\langle\,,\rangle$ is the Frobenius inner product defined as $\langle\matA,\matB\rangle = \mattr(\matB'\matA)$, and $\eta$ is the adaptive penalty parameter.
Now, the above equation can be equivalently re-written as (disregarding the constant terms)
\small
\begin{align}
\matP_{j+1} = \argmin_{\matP\in\calD} \frac{1}{2}\|\matP - \big(\matP_j + \tau\matZ'(\matM-\matZ\matP)\big)\|^2
\label{eq:proj_op1}
\end{align}
\normalsize
where $\matM = \bomega - \bsigma$.
Note that we do not introduce any Lagrangian multipliers corresponding to the auxiliary variables $\bomega$ and $\bsigma$.
By initializing $\bomega$ to $\balpha$ and $\bsigma$ to $\bbeta$, the Lagrangian variables $\matU$ and $\matV$ stand as the Lagrangian variables for $\bomega$ and $\bsigma$ as well. In a similar fashion, the second and third terms in the line \ref{alg:3line} can be equivalently re-written as:
\small
\begin{align}
\matP_{j+1} = \argmin_{\matP\in\calD} \frac{1}{2}\|\matP - \big(\matP_j + \tau(\matN-\matP\matX)\matX'\big)\|^2
\label{eq:proj_op2}
\end{align}
\normalsize
where $\matN = \balpha+\matU$, and
\small
\begin{align}
\matP_{j+1} = \argmin_{\matP\in\calD} \frac{1}{2}\|\matP - \big(\matP_j + \tau\matY'(\matO-\matY\matP)\big)\|^2
\label{eq:proj_op3}
\end{align}
\normalsize
where $\matO = \bbeta + \matV$

Now, combining Equations \eqref{eq:proj_op1},\eqref{eq:proj_op2},\eqref{eq:proj_op3}, our final optimization problem for $\matP$ becomes,
\small
\begin{align}
\matP_{j+1} &= \argmin_{\matP\in\calD} \frac{1}{2}\Big\|\matP - \frac{\big(\matA + \matB + \matC)}{3}\Big\|^2 
\label{eq:proj_op}
\end{align}
\normalsize
where $\matA = \matP_j + \tau\matZ'(\matM-\matZ\matP), \matB = \matP_j + \tau(\matN-\matP\matX)\matX'$ and $\matC = \matP_j + \tau\matY'(\matO-\matY\matP)$.

Equation \eqref{eq:proj_op} corresponds exactly to the euclidean projection of the matrix $\nicefrac{(\matA+\matB+\matC)}{3}$ onto the set of double stochastic matrix.
The problem of projecting a matrix onto the Birkhoff polytope is a well studied problem, and there exists very simple and efficient algorithm as shown in \citep{sinkhorn1967concerning}.
The algorithm proceeds by alternately normalizing rows and columns of the given non-negative matrix.
In fact, it is established that Sinkhorn projection algorithm returns a double stochastic matrix which is optimal according to KL-divergence i.e. the double stochastic matrix has the lowest KL-divergence distance rather than euclidean distance.
A recent paper \citep{wang2010learning} proposed a projection algorithm which returns a projection which is optimal with respect to the euclidean projection.
We use the algorithm proposed in \cite{wang2010learning} to project $\matP$ onto the Birkhoff polytope.

\subsubsection{Solving for $\matZ$}
The subproblem at line \ref{alg:4line} corresponding to $\matZ$ is also a constrained optimization problem.
Unlike for $\matP$, here we can find a simple closed-form solution for the projection.
The problem is to find the closest projection of a given matrix $\matZ$ to the set of the matrices $\calF$.
The set of matrices $\calF$ is defined as the set of all matrices whose entries lie in the interval [0,1].
The problem can be written as ${\displaystyle \matZ^{*}=\arg\min_{\matA\in\calF}||\matA-\matZ||^2}$.
This is a convex problem, where the objective becomes ${ \displaystyle \|\matA-\matZ\|^2=\sum_{i,j}(\matA_{ij}-\matZ_{ij})^2}$ and thus, it is enough to find solutions for $\matA_{ij}$ individually.
\small
\begin{equation}
\matA_{ij}=\begin{cases}
\matZ_{ij}, & \mbox{if}~0<\matZ_{ij}<1\\
0, & \mbox{if}~-\infty<\matZ_{ij}<0 \\
1, & \mbox{if}~~1<\matZ_{ij}<\infty
\end{cases}
\label{eq:z_proj}
\end{equation}
\normalsize
We solve for $\matZ$ using the projected gradient descent method.
The gradient of $\matZ$ at the point $\matZ_j$ can be estimated as,
\small
\begin{equation*}
\bnabla\matZ\big|_{\matZ_j} =  0.5*(\matZ_j*\matP-(\bbeta-\balpha))\matP' + \nu(\matZ_j-\matD-\matC)
\end{equation*}
\normalsize
At the end of each gradient descent step, we project the result to the set of all matrices whose entries lie in the interval [0,1], using the projection rule defined in Equation \ref{eq:z_proj}.
We project the resulting ${\matZ}^{*}$ to the set of symmetric matrix having minimum euclidean distance by using
$\Pi(\matZ^{*}) = \frac{{\matZ}^{*}+{\matZ}^{*}{'}}{2}$.
\section{Experiments \& Analysis}
\label{sec:exp}
We evaluate the performance of our algorithm on several synthetic and real world datasets.
The objective of our experiments is three-fold: 1) including the structured perturbation helps reducing the adjacency mismatches thus obtaining lower matching error; 2) one can infer the structured perturbations with high precision and recall in simulated networks where the true $\matZ$ is known; and 3) evaluate the structured perturbations as significant clusters i.e. block-diagonals in a node-cluster membership based re-ordered $\hat{\matZ}$ using the unsupervised F-measure defined as the harmonic mean of entropy and balance 
\cite{mall2013self}.

The probability for cluster $i$ is defined as: $p_{i} = \nicefrac{|C_{i}|}{n}$, where $C_{i}$ represents the nodes in $\hat{\matZ}$ belonging to structured perturbation cluster $i$.
Entropy is defined as: $E = \sum_{i=1}^{k} p_{i}\log(p_{i})$, 
which ranges from $0$, when all the $n$ nodes are part of an inferred structured perturbation, to a maximum of $\log(n)$, when each individual node is a structured perturbation cluster.
Normalized entropy is defined as: $NE = \nicefrac{E}{\log(n)}$ to make its range $[0,1]$.
Similarly, balance is defined as: $B = \nicefrac{(\sum_{i}^{k} \frac{|C_{i}|}{\max(|C_1|,|C_2|,\ldots,|C_k|)})}{k}$.
Balance values are high when a few large clusters are identified as structured perturbation in the inferred $\hat{\matZ}$.
Thus, there exist a trade-off between $E$ and $B$.
Our goal is to identify the $k$ that yields the optimal unsupervised F-measure.
This gives a good way to measure the performance on real paired networks where true $\matZ$ is unknown.
Further details about using F-measure can be found in \cite{mall2013self}.

In the first set of experiments, following \cite{fiori2013robust}, we compare the matching error against state-of-the-art graph matching algorithms on different synthetic graphs. We used the algorithm UNMY \citep{umeyama1988eigendecomposition}, IsoRank (RANK) \cite{singh2007pairwise}, PATH \cite{zaslavskiy2009path} and multi-modal graph matching (MMG) \cite{fiori2013robust} as our baselines.
Our proposed algorithm is denoted as STEPD. In STEPD, the regularization co-efficients ($\nu$ and $\mu$) are selected from the range $\{2^{-5},\cdots ,1\}$ as multiple of 2 that resulted in the best matching error.
It has been shown that ADMM works well without any tuning of the penalty parameter $\rho$, and a value of 1 works well in most settings \cite{boyd2011distributed}.
The number of ADMM iteration is set to 150. The matching error for STEPD is defined as $\|\matP\matX + \matZ\matP-\matY\matP\|$, as $\matZ$ is not noise whereas in baselines it is defined as $\|\matP\matX - \matY\matP\|$.
\subsection{ Data}
We experimented with $5$ randomized 
Scale-Free (SF) graphs (with exponent $\alpha$=$1.5$) and Erdos-R$\grave{e}$nyi graphs (ER) (with parameter $r$=$0.15$) respectively and the reported results are the average over the $5$ runs.
For each experiment, the number of vertices ($n$) was set to $500$, two structured perturbations (by adding edges to form cliques) were introduced in $\matY$, one of $50$ nodes while the other of $100$ nodes.
We also added an additional $30$ or $50$ random edges in $\matY$ as noise. We had $2$ set of experiments: 1) the permutation matrix ($\matP$) was set to an identity matrix ($\matI$) referred as `NP' and 2) the first $50$ nodes of $\matX$ were permuted referred as `P'. Table \ref{table:ADME} shows a comprehensive comparison of STEPD with state-of-the-art graph matching algorithms 
w.r.t. matching error. Supplementary Figure 1 
depicts the effect of STEPD model parameters $\nu$ and $\mu$ on 
precision, recall and F-score, when comparing the structured perturbations in the original $\matZ$ with the inferred $\hat{\matZ}$ for both SF and ER graphs.
\begin{table}
    \setlength{\tabcolsep}{2.8pt}
    \centering
    \caption{Synthetic Data Mean Matching Error }
    \label{table:ADME}
    \begin{tabular}{l*{6}{c}}
        \hline \\
        Graph      & \#noise &  UMY  &  RANK  &  PATH &  MMG  &  STEPD \\ \hline
        SF (NP)    &    30   & 84.07 & 85.01 & 84.24 & 82.60 & 65.19 \\
        SF (NP)    &    50   & 84.85 & 85.78 & 82.33 & 82.64 & 66.22 \\
        SF (P)     &    30   & 84.34 & 85.51 & 84.08 & 81.72 & 60.29 \\
        SF (P)     &    50   & 85.05 & 85.58 & 77.71 & 82.42 & 60.34 \\
        ER (NP)    &    30   &172.18 &173.33 &103.33 &134.55 &118.27 \\
        ER (NP)    &    50   &172.49 &173.24 & 83.31 &125.14 &103.54 \\ 
        ER (P)     &    30   &172.19 &173.29 & 75.13 &121.99 &114.39 \\
        ER (P)     &    50   &170.96 &172.54 & 83.80 &122.25 &114.81 \\
        \hline
    \end{tabular}
\end{table}

\subsection{Spectral Recovery Evaluation}
\begin{figure}[htbp]
   \centering
    \centerline{\includegraphics[width=0.80\linewidth, height=3.5cm]{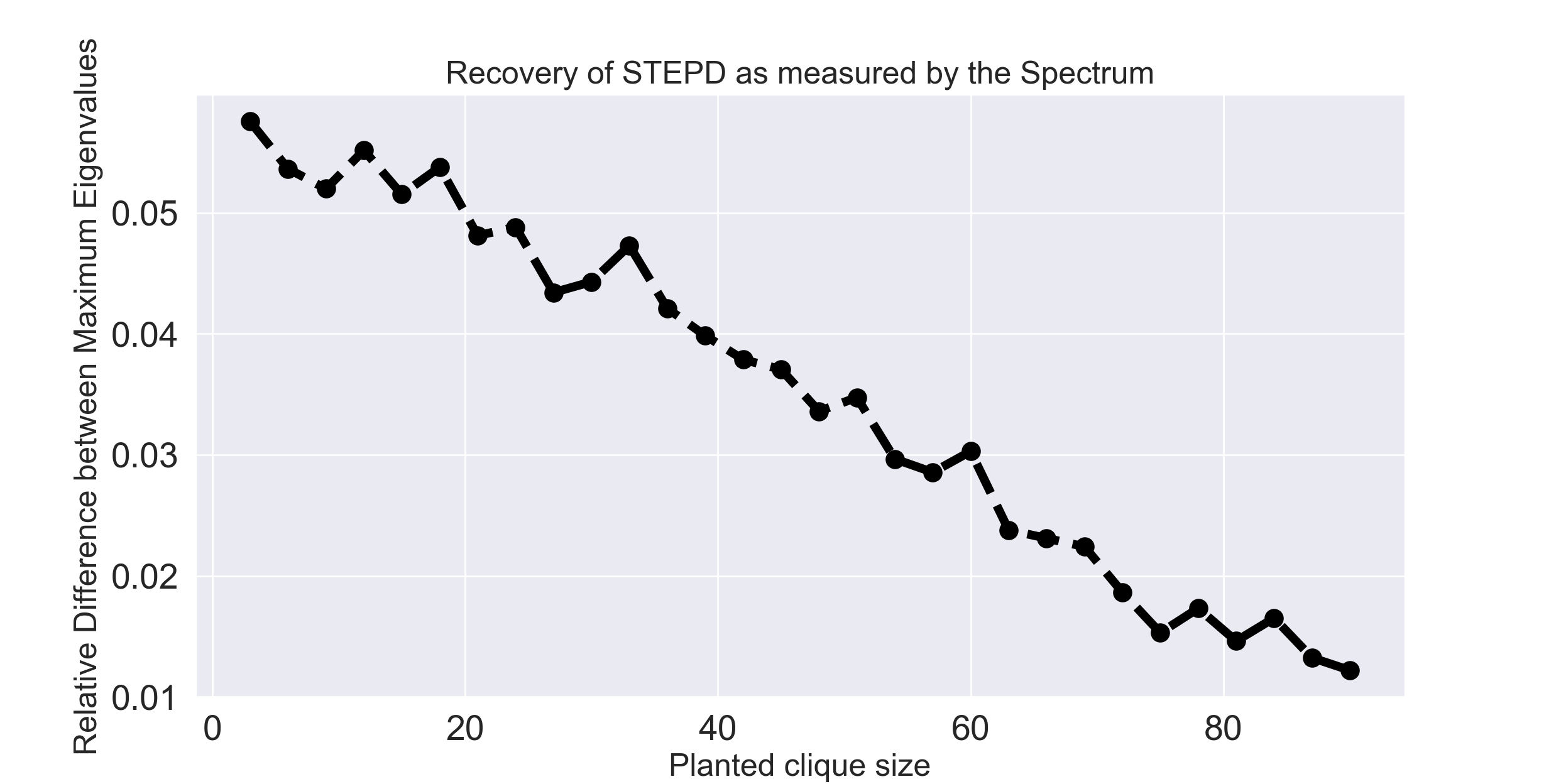}}
    \caption{Relative difference between the maximum eigenvalues between $Y$ and $\hat{Y}$}

\label{fig:spectrum}
\end{figure}

We demonstrated the ability of STEPD to recover permuted planted cliques in the presence of bernoulli noise with Synthetic data. We generated Erodos-Reyni (ER) graph $\matX$ of size $100$ and probability $0.5$. We then generated permuted graphs $\matP$ where five nodes were randomly permuted and the others
were kept fixed. We then planted cliques of sizes varying as multiples of three up to a maximum clique of size $90$. Furthermore, we added noisy random edges again, using an ER graph of size 100 and probability 0.1. Thus, the permuted and perturbed matrix was formed as $\matY = \matP\matX\matP' + \matZ + \matE$. STEPD was given input $\matX$ and $\matY$ and output two matrices $\hat{\matP}$ and $\hat{\matZ}$, the inferred permutation and the structured perturbation matrices. We then generated
the matrix $\hat{\matY}=\hat{\matP}\matX\hat{\matP'}+\hat{\matZ}$ and computed the relative
change in the maximum eigenvalues of $\hat{\matY}$ and $\matY$ defined as $\frac{\lambda_{max}(\hat{\matY}) - \lambda_{max}(\matY)}{\lambda_{max}(\matY)}$. The result of the relative change as a function of the planted clique size are shown in Figure~\ref{fig:spectrum}. It is clear that as the size of the clique increases, the ability of STEPD to recover matrices which are spectrally equivalent to the original $\matY$ increases. Moreover, even for small planted cliques the relative error is small. 

\vspace{-0.4cm}

\subsection{Real Data}
We compare the matching errors of different algorithms on two real world data.
\subsubsection{C. Elegans network}
\begin{figure*}[t]
\centering
\includegraphics[width=0.85\linewidth]{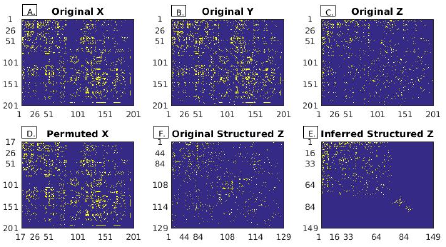}
\caption{An example of two roundworm (C. Elegans) networks demonstrating the PSPI problem and solution. Each subplot shows the adjacency matrix of the respective graph. Here the `yellow' color represents the non-zero edges in the adjacency matrix. STEPD identifies $4$ structured perturbations as block diagonal components in the inferred $\hat{\matZ}$ with a precision of $0.703$ and recall of $0.626$ w.r.t. the structured perturbations in the original $\matZ$.}
\label{fig:fig1}
\end{figure*}

An experiment was performed on the neuronal connectivity networks of two roundworms belong to the C. Elegans family \cite{white1986structure}. 
The nodes of the network are a common set of $202$ neurons with different number of edges, $2870$ and $3090$ respectively. The adjacency
matrices $\mathbf{X}$ and $\mathbf{Y}$ are shown in Figure~\ref{fig:fig1}.


As shown in \cite{fiori2013robust}, we permuted the first $20$ nodes of $\matX$ to obtain $\matX_p$ s.t. $\matY = \matP\matX_p\matP' + \matZ$ and our goal is to infer both $\matP$ and $\hat{\matZ} \sim \matZ$. 
Figure \ref{fig:fig1} shows a node-cluster membership based re-ordered version of $\matZ$ to reflect the structured perturbations in $\matZ$ as approximate block diagonals. The corresponding structured perturbations in the inferred $\hat{\matZ}$ obtained by the proposed STEPD methodology is shown in Figure~\ref{fig:fig1}F. Table \ref{table:t2} compares the matching error of STEPD algorithm with several graph matching approaches. The STEPD algorithm has the least matching error. 

Moreover, we perform our unsupervised F-measure based analysis to identify structured perturbations (i.e. clusters) in both $\matZ$ and inferred $\hat{\matZ}$. The optimal $k$ for which F-measure was maximum for $\matZ$ was $k=4$ with F-measure$=$0.33. Similarly, the optimal $k$ for which F-measure was maximum for $\hat{\matZ}$ was $k=4$ with F-measure$=$0.35. This suggests that our STEPD algorithm correctly identifies the structured perturbation clusters in $\matZ$. To have a comprehensive analysis, we further compared the location of non-zero weight edges i.e. local structure of the perturbation clusters in $\matZ$ and inferred $\hat{\matZ}$ to attain a precision of $0.703$ and recall of $0.626$.
\begin{table}
    \centering
    \caption{Real Data Matching Error }
    \label{table:t2}
    \begin{tabular}{l*{5}{c}}
        \hline \\
        Graph    &  UMY  &  RANK  & PATH  &  MMG  &  STEPD \\ \hline
        CEL      & 71.57 & 71.36 & 65.18 & 64.25 & 57.98 \\
        TCGA     & 124.97 &122.70 &141.89 &113.50 & 109.82 \\
        \hline
    \end{tabular}
\end{table}
\subsubsection{Glioma Cancer Data}
We analyze the structures of two networks extracted from a glioma cancer
dataset obtained from the Cancer Genome Atlas (TCGA)\cite{johnson2007adjusting}. The two networks correspond to IDH-mutant and IDH-wildtype subtypes of glioma and
are labeled $\mathbf{X}$ and $\mathbf{Y}$ respectively. The networks includes $12,985$ genes comprising $457$ transcription factors (TF) and $12,895$ target (T) genes, where $\text{TF} \in \text{T}$. The IDH-mutant network has $13,683$ (TF $\rightarrow$ T) edges while IDH-wildtype network has $14,158$ (TF $\rightarrow$ T) edges. Our goal is to identify sub-networks of TFs having a different regulatory program in these glioma subtypes.

\begin{figure*}[!ht]
	\centering
    \begin{subfigure}{0.5\textwidth}
    	\centering
        \includegraphics[width=0.9\linewidth]{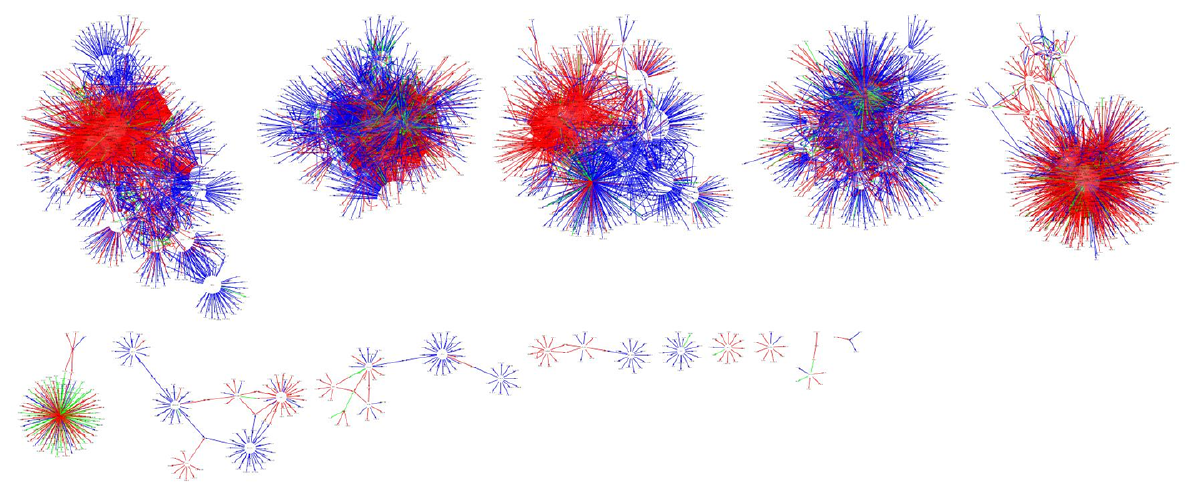}
        \caption{Original sub-network of $286$ TFs which are part of  structured perturbation cluster $C_1$.}
        \label{fig3:subfig1}
    \end{subfigure}
    \begin{subfigure}{0.2\textwidth}
    	\centering
        \includegraphics[width=0.9\linewidth]{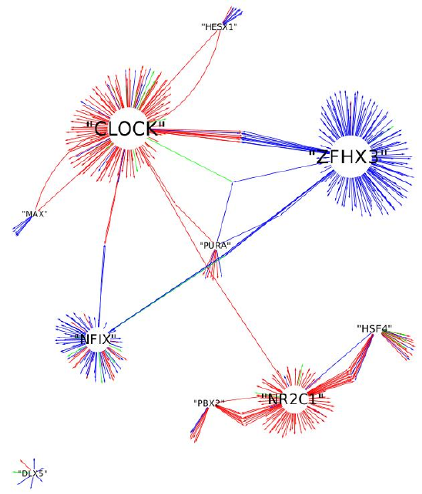}
        \caption{Original sub-network of $10$ TFs present in structured perturbation cluster $C_2$.}
        \label{fig3:subfig2}
    \end{subfigure}
    \begin{subfigure}{0.23\textwidth}
    	\centering
        \includegraphics[width=0.9\linewidth]{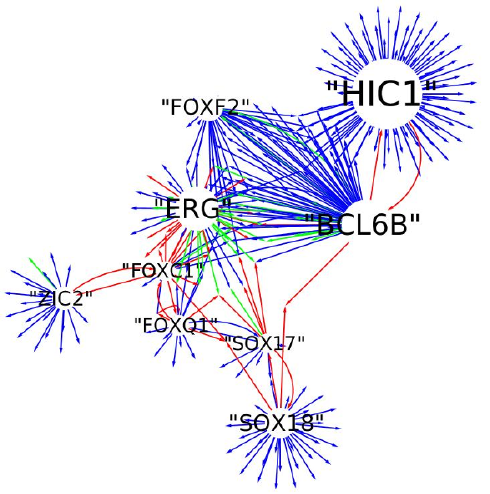}
        \caption{Original sub-network of $10$ TFs present in structured perturbation cluster $C_3$.}
        \label{fig3:subfig3}
    \end{subfigure}
    \caption{Original sub-networks in the IDH-mutant and IDH-wildtype graphs consisting of the edges belonging to the 3 perturbation clusters  in $\hat{\matZ}$ inferred by STEPD algorithm. Here `green' edges represent edges present in both IDH-mutant and IDH-wildtype. The `red' edges are from the IDH-mutant graph while `blue' edges come from IDH-wildtype network. The `white' circles represent the TFs and the peripheral nodes are the target (T) genes. In accordance with our goal of identification of difference in regulatory networks of IDH-mutant versus IDH-wildtype, we observe that both perturbation clusters $C_1$ and $C_2$ have distinct communities (predominatly `red' or `blue' connections) associated with IDH-mutant and IDH-wildtype sub-networks respectively. 
Thus, STEPD algorithm can detect structured changes between IDH-mutant and IDH-wildtype sub-networks which have an equivalent biological relevance. We present the detail of the unveiled relevance in Table.2 of the Supplementary.}
    \label{fig:fig3}
\end{figure*}


Though there exists one-to-one mapping between the nodes in $\matX$ and $\matY$, details about the inferred permutation matrix and genes taking over role of other genes is provided in Supplementary. Here we focus on structured perturbation. Using the F-measure criterion \cite{mall2013self}, we detected optimal $k=3$ at F-measure$=$0.30 for $\matZ$. Similarly, we identified $3$ structured perturbations in the $\hat{\matZ}$ using STEPD method at F-measure$=$0.30 (for $\nu=0.5$,$\mu=0.5$). We attained a  precision of $1.0$ and recall of $0.952$ when comparing the edges in the structured perturbations of $\matZ$ and $\hat{\matZ}$. The largest cluster comprised of $286$ genes (TFs), while the $2$ smaller clusters consisted of $10$ TFs each. We showcase the original sub-networks of the IDH-mutant and IDH-wildtype graphs corresponding to the TFs in each cluster in Figure \ref{fig:fig3}. Additional caveats are provided in Supplementary.

To carry out a deeper biological analysis, we use the GO Onotology (GO terms) 
which comprises a repository of known functions and processes associated with
genes. For example, each gene can be classified as participating in
a Biological Process (BP), Cellular component (CC), Molecular Function (MF)
or Pathways (PW). Each of these high level processes and functions 
consist of terms, for example, ``defense response to virus'' or ``protein
modification process.'' For each perturbation cluster, we identified the enriched
or over-represented GO terms using ConsensusPathDB \cite{kamburov2012consensuspathdb}.

A detailed breakdown of the enriched GO terms for both the
IDH-mutant ($\mathbf{X}$) and IDH-wildtype ($\mathbf{Y}$) case is provided in Supplementary. It was shown in \cite{ceccarelli2016} that the main difference between IDH-mutant and IDH-wildtype gliomas is the characteristic hyper-methylation phenotype i.e. chromatin modification and histone acetylation. Furthermore, it was shown in \cite{frattini2018metabolic} that \textbf{PPAR$\alpha$} is recruited for mitochondrial respiration leading to tumor oncogenesis in fusion gliomas. This is indicated by the over-represented biological components and pathways for IDH-mutant and IDH-wildtype case respectively for perturbation cluster $C_2$. Details of the enriched biological components and pathways associated with perturbation clusters $C_1$ and $C_2$ are provided in the Supplementary. The STEPD approach is not only able to identify known but also detects potential novel enrichments which need further investigation. 

\vspace{-2mm}
\section{Conclusion}
In this paper, we introduced the problem of Permutation and Structured Perturbation Inference (PSPI), a combinatorial problem with real world applications in the fields of system biology and computer vision. PSPI is a generalization of the graph matching problem and takes structured perturbations into account while computing the permutations between a pair of graphs.
We proposed {\tt STEPD}, an iterative algorithm to solve the relaxed version of the problem.
Experiments on computational biology benchmark datasets showed that PSPI can be used to make potentially insightful discoveries from gene regulatory networks  and can become an important prognostic tool for biologists. Source code and datasets for reproducibility can be found at https://github.com/code-halo/rgm/.

\bibliography{paper_v2.bib}

\end{document}